\documentclass[sigconf]{acmart}

\AtBeginDocument{%
  }
\usepackage{multirow}
\usepackage[ruled,vlined,linesnumbered]{algorithm2e}

\copyrightyear{2025}
\acmYear{2025}
\setcopyright{rightsretained}
\acmConference[RecSys '25]{Proceedings of the Nineteenth ACM Conference on Recommender Systems}{September 22--26, 2025}{Prague, Czech Republic}
\acmBooktitle{Proceedings of the Nineteenth ACM Conference on Recommender Systems (RecSys '25), September 22--26, 2025, Prague, Czech Republic}\acmDOI{10.1145/3705328.3759312}
\acmISBN{979-8-4007-1364-4/2025/09}

\begin{document}

\title{RicciFlowRec: A Geometric Root Cause Recommender Using Ricci Curvature on Financial Graphs}


\author{Zhongtian Sun}
\affiliation{%
  \institution{University of Kent}
  \city{Canterbury}
  \country{UK}}
\email{z.sun-256@kent.ac.uk}

\author{Anoushka Harit}
\affiliation{%
  \institution{University of Cambridge}
  \city{Cambridge}
  \country{UK}}
\email{ah2415@cam.ac.uk}


\renewcommand{\shortauthors}{Sun et al.}

\begin{abstract}
We propose \textbf{RicciFlowRec}, a geometric recommendation framework that performs root cause attribution via Ricci curvature and flow on dynamic financial graphs. By modelling evolving interactions among stocks, macroeconomic indicators, and news, we quantify local stress using discrete Ricci curvature and trace shock propagation via Ricci flow. Curvature gradients reveal causal substructures, informing a structural risk-aware ranking function. Preliminary results on S\&P~500 data with FinBERT-based sentiment show improved robustness and interpretability under synthetic perturbations. This ongoing work supports curvature-based attribution and early-stage risk-aware ranking, with plans for portfolio optimization and return forecasting. To our knowledge, RicciFlowRec is the first recommender to apply geometric flow-based reasoning in financial decision support.
\end{abstract}



\maketitle

\section{Introduction}
Recommendation systems ~\cite{sun2025advanced,harit2025manifoldmind} are increasingly used in financial domains, supporting tasks such as asset selection and portfolio construction~\cite{fischer2018deep,chen2015lstm}. However, many existing systems rely on correlation-based signals or opaque neural models, which are vulnerable to market shocks and offer limited interpretability~\cite{zhang2020explainable}. Recent advances in causal inference and logic ~\cite{bonner2018causal,harit2024monitoring,sun2025actionable,harit2025causal,sun2025glance}, temporal graph learning~\cite{sun2023money,lee2024stock,sun2025spar} and knowledge graphs~\cite{harit2024breaking} have improved structural awareness, yet often overlook the geometric organisation of financial markets. Existing models fail to capture how localized shocks propagate through interdependencies and affect distant assets. We introduce \textbf{RicciFlowRec}, a novel recommendation framework that applies geometric root cause analysis (RCA) on dynamic financial graphs. It uses discrete Ricci curvature ~\cite{ni2019community} to quantify structural tension and simulates Ricci flow ~\cite{ni2015ricci,sun2023rewiring} to track stress evolution. Nodes and subgraphs with large curvature changes are identified as root causes of instability, informing a curvature-aware ranking function that penalizes exposure to unstable regions.



Our graph includes stocks, macroeconomic indicators, and news events, with edges based on co-movement, semantic similarity (FinBERT~\cite{araci2019finbert}) and sector relationships. The curvature-informed ranking adjusts predicted returns based on structural risk, ensuring robust and interpretable recommendations. RicciFlowRec supports curvature computation, RCA path tracing and stress-aware scoring. Preliminary results on S\&P 500 data show RicciFlowRec improves ranking stability under synthetic shocks and highlights key stress paths.\textbf{Our main contributions are as follows:}
\begin{itemize}
    \item We propose \textbf{RicciFlowRec}, a curvature-aware recommendation framework that performs geometric root cause analysis on financial graphs.
    
    \item We introduce a ranking method that adjusts return predictions based on each asset’s exposure to structurally unstable regions identified via Ricci flow.
    
    \item We implement a prototype supporting curvature computation, stress path tracing, and initial asset re-ranking.
    
    \item We present results showing improved ranking stability and interpretability on S\&P 500 data with FinBERT sentiment.
\end{itemize}
RicciFlowRec provides a principled, interpretable approach to financial recommendations, boosting robustness and transparency under market volatility.

\section{Problem Formulation}
Let $\mathcal{A} = \{a_1, a_2, \dots, a_N\}$ denote a universe of $N$ financial assets (e.g., equities or ETFs). At each time $t$, each asset $a_i \in \mathcal{A}$ is associated with a feature vector $\mathbf{x}_i^t \in \mathbb{R}^d$, capturing historical returns, sentiment scores, and sector or macroeconomic attributes. We model the financial system as a dynamic heterogeneous graph:
\[
\mathcal{G}_t = (\mathcal{V}_t, \mathcal{E}_t, \mathbf{W}_t)
\]
where:
\begin{itemize}
    \item $\mathcal{V}_t$: nodes include assets, macroeconomic indicators, and news entities;
    \item $\mathcal{E}_t$: edges encode relationships such as return correlations, semantic similarity, and economic dependencies;
    \item $\mathbf{W}_t$: weights represent the interaction strength of each edge at time $t$.
\end{itemize}

The objective is to recommend a subset of $K$ assets at time $t$:
\[
\mathcal{R}_t = \left[ a_{(1)}, \dots, a_{(K)} \right] = \arg\max_{\substack{\mathcal{S} \subseteq \mathcal{A} \\ |\mathcal{S}| = K}} \sum_{a_i \in \mathcal{S}} U(a_i, \mathcal{G}_t)
\]
where $U(a_i, \mathcal{G}_t)$ integrates:
\begin{itemize}
    \item $\hat{r}_i^{t+\tau}$: expected return over a future horizon $\tau$;
    \item $\rho(a_i, \mathcal{G}_t)$: geometric risk signal reflecting structural stress exposure.
\end{itemize}

In practice, the scoring function $s(a_i)$ is the operational instantiation of $U(a_i, \mathcal{G}_t)$:
\begin{equation}
s(a_i) = \alpha \cdot \hat{r}(a_i) - (1-\alpha) \cdot \mathrm{Risk}(a_i),
\end{equation}
where $\hat{r}(a_i)$ is the predicted return and $\mathrm{Risk}(a_i)$ denotes the normalised exposure of asset $a_i$ to unstable or dynamically shifting regions in $\mathcal{G}_t$.

\begin{figure}[h]
    \centering
    \includegraphics[width=0.55\linewidth]{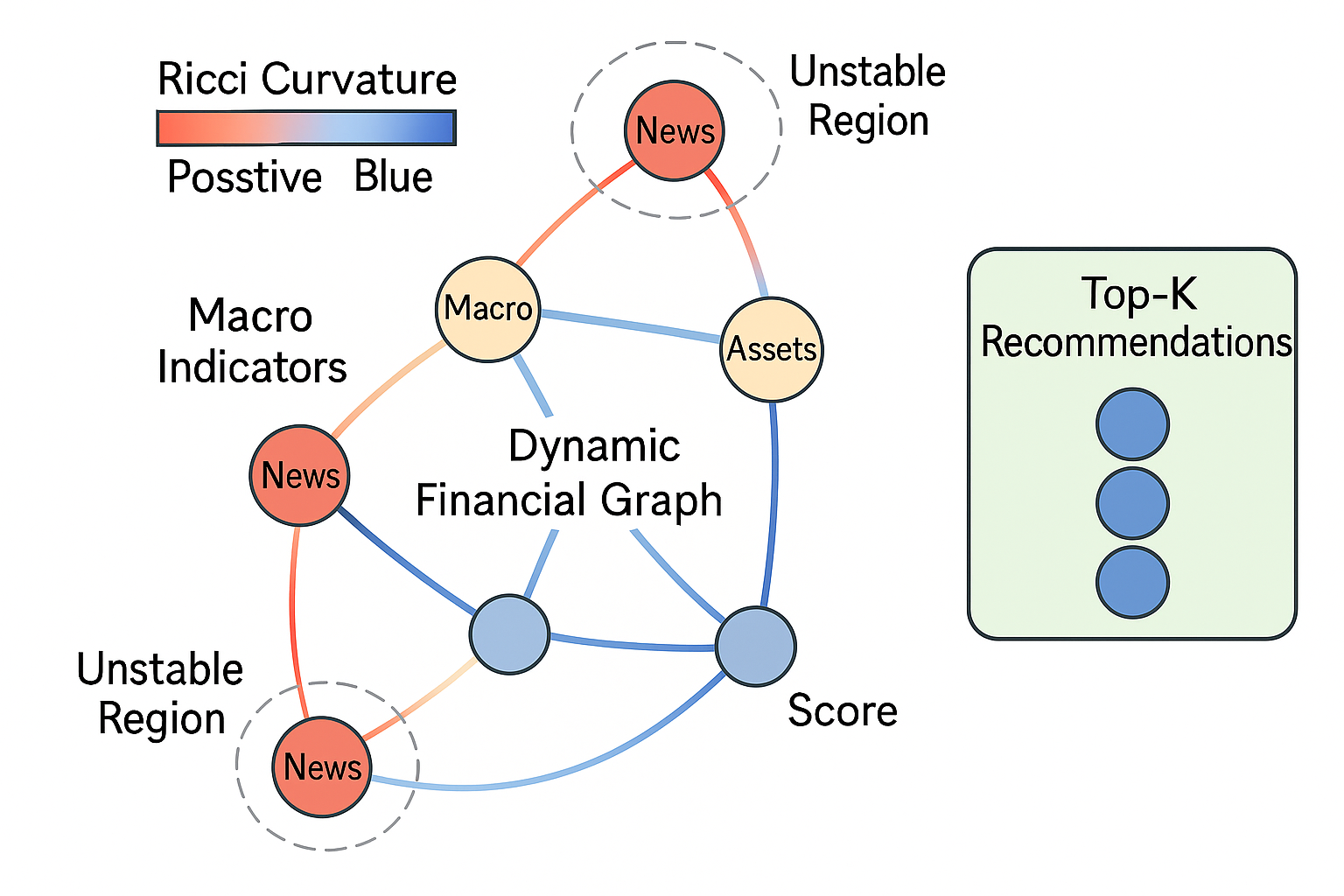}
    \caption{Dynamic financial graph $\mathcal{G}_t$ with Ricci curvature $\kappa(e)$ shown on edges (red: unstable, blue: stable). Nodes represent assets, macro indicators, and news entities. Unstable zones inform $\rho(a_i, \mathcal{G}_t)$ for scoring and Top-$K$ recommendations.}
    \label{fig:ricci_graph}
\end{figure}

\textbf{Challenges.}
\begin{enumerate}
    \item Structural risk emerges from long-range dependencies that are not captured by local neighbourhoods.
    \item Most systems fail to attribute volatility to root causes in the financial graph.
    \item High-return assets may exhibit latent geometric instability, making them vulnerable to cascading failures when market structure shifts.
\end{enumerate}

\textbf{Objective.}
We define a scoring function:
\[
s(a_i) = f\left( \hat{r}_i^{t+\tau}, \rho(a_i, \mathcal{G}_t) \right)
\]
where $\rho(a_i, \mathcal{G}_t)$ quantifies how much $a_i$ is exposed to unstable or dynamically shifting regions in the financial graph. Our proposed model, \textbf{RicciFlowRec}, integrates discrete Ricci curvature and simulated Ricci flow to detect root causes of market stress and guide robust, interpretable asset selection.

\section{Methodology}
\textbf{RicciFlowRec} is a curvature-aware recommendation framework designed to detect latent sources of instability in financial markets and incorporate structural risk into asset ranking. The system integrates discrete Ricci curvature and Ricci flow into a dynamic graph neural architecture, enabling root cause attribution (RCA) and robust decision-making. 

\begin{figure}[htbp]
  \centering
  \includegraphics[width=0.6\linewidth]{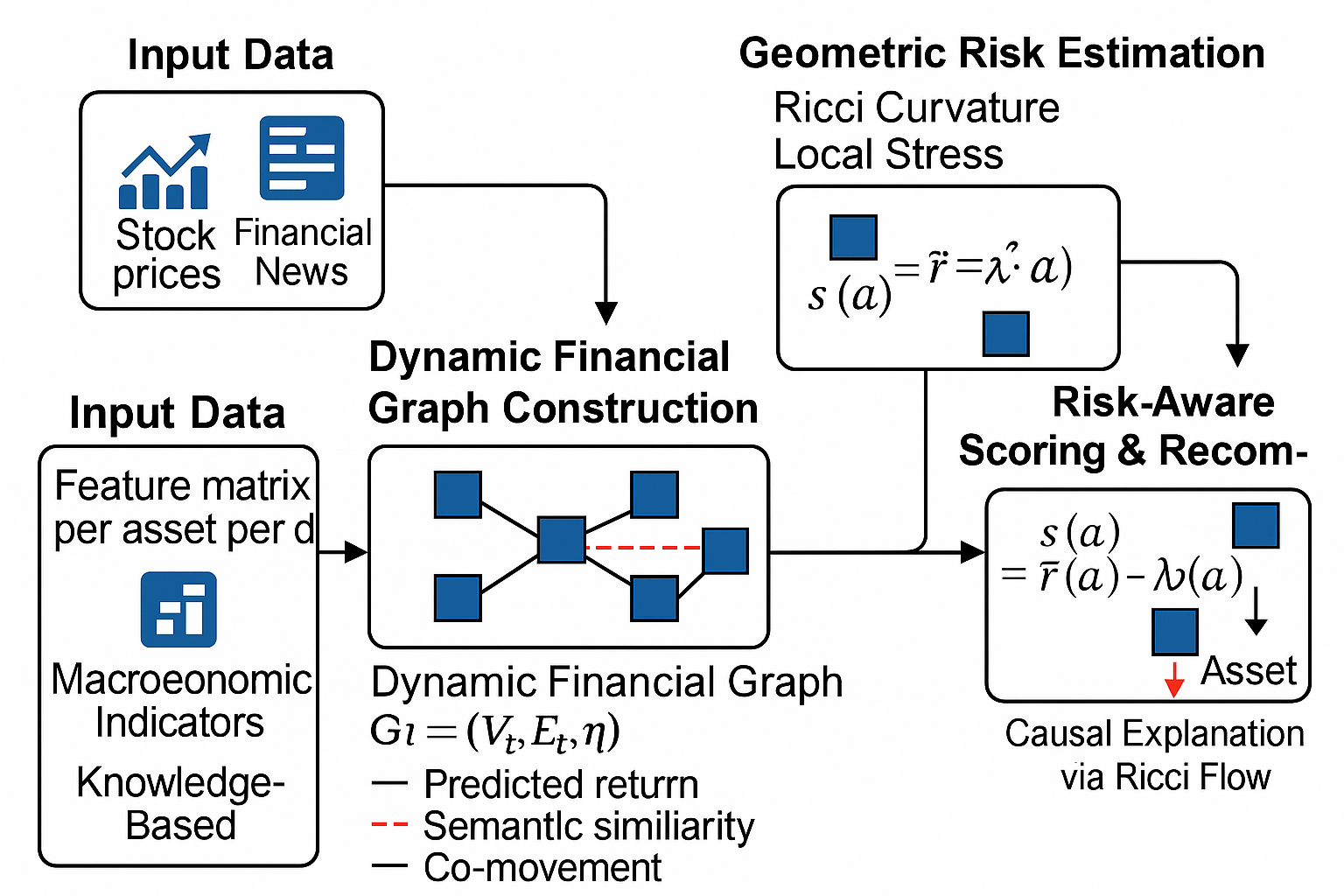} 
  \caption{Overview of the RicciFlowRec pipeline. Input data are fused into a dynamic graph, where Ricci curvature and flow inform risk-aware asset scoring.}
  \label{fig:pipeline}
\end{figure}

\subsection{Graph Construction from Market and News Signals}
At each trading day $t$, we construct a heterogeneous financial graph $\mathcal{G}_t = (\mathcal{V}_t, \mathcal{E}_t, \mathbf{W}_t)$ capturing dynamic relations among financial entities. The node set $\mathcal{V}_t$ comprises:
\begin{itemize}
    \item \textbf{Assets:} S\&P 500 equities;
    \item \textbf{Macroeconomic indicators:} e.g., bond yields, CPI;
    \item \textbf{News entities:} Named entities and headlines extracted and embedded via FinBERT~\cite{araci2019finbert}.
\end{itemize}

Edges $(u, v) \in \mathcal{E}_t$ are constructed using three signals:
\begin{enumerate}
    \item \textbf{Rolling correlations:} Pearson correlation of log returns over a 30-day window;
    \item \textbf{Semantic proximity:} Cosine similarity between FinBERT embeddings of co-mentioned news;
    \item \textbf{Knowledge links:} Sectoral or supply-chain relations from external economic graphs~\cite{harit2024breaking}.
\end{enumerate}
The edge weights $w_{uv}^t$ reflect interaction strength and serve as input for curvature estimation. These graphs are updated daily to reflect evolving market and information structures.

\subsection{Discrete Ricci Curvature for Stress Detection}
We compute Ollivier-Ricci curvature~\cite{ni2019community} to characterise edge-level geometry and detect local structural fragility. For each edge $(u, v) \in \mathcal{E}_t$:
\[
\kappa(u,v) = 1 - \frac{W_1(\mu_u, \mu_v)}{d(u,v)}
\]
where $W_1$ is the Wasserstein-1 distance between node neighbourhood distributions $\mu_u$, $\mu_v$, and $d(u,v)$ is the graph distance. Interpretation:
\begin{itemize}
    \item \textbf{Negative curvature:} Structural divergence or bottleneck, indicates potential root causes of systemic stress;
    \item \textbf{Positive curvature:} Redundant or tightly coupled structures — suggests robustness.
\end{itemize}

This curvature landscape provides a topological signature of systemic health and is recomputed as the graph evolves.

\subsection{Simulating Ricci Flow for Root Cause Attribution}
To identify dynamic sources of stress, we simulate Ricci flow over a 5-day horizon. The curvature-adjusted edge weights evolve as:
\[
\frac{d}{dt} w_{uv}(t) = -\kappa(u,v) \cdot w_{uv}(t)
\]
This evolution penalises fragile connections and highlights emerging vulnerabilities. We define the curvature shift:
\[
\Delta \kappa(u,v) = \kappa_{t+\Delta t}(u,v) - \kappa_t(u,v)
\]
Edges with large $|\Delta \kappa|$ signal temporal instability. We extract \textbf{RCA paths} by tracing high-magnitude curvature shifts backward through $\mathcal{G}_t$, surfacing subgraphs that explain risk propagation.

\subsection{Structural Risk Scoring for Recommendation}
Let $\hat{r}_i^{(t+\tau)}$ be the predicted return of asset $a_i$ over horizon $\tau$ from a base forecaster (e.g., LSTM, GAT). We define an RCA-aware adjustment:
\[
s(a_i) = \hat{r}_i^{(t+\tau)} - \lambda \cdot \rho(a_i)
\]
where
\[
\rho(a_i) = \sum_{v \in \mathcal{N}(a_i)} |\Delta \kappa(a_i, v)|
\]
is the \textbf{structural exposure} of $a_i$ to curvature perturbations, and $\lambda > 0$ controls the risk sensitivity. Assets heavily connected to volatile regions are down-weighted. This scoring function promotes stability-aware rankings.

\subsection{Final Recommendation and Explanation}
The final recommendation set is:
\[
\mathcal{R}_t = \text{TopK}_{a_i \in \mathcal{A}}(s(a_i))
\]
Each recommended asset is accompanied by an \textbf{RCA subgraph}, a path-traced explanation highlighting the structural drivers of ranking adjustment. This enables transparent decision support for investors and analysts.

\subsection{Design Highlights}
RicciFlowRec offers three core advantages:
\begin{enumerate}
    \item \textbf{Causal Geometric Attribution:} Quantifies and localises the source of stress using curvature dynamics;
    \item \textbf{Robust Recommendation:} Penalises structurally exposed nodes to improve resilience under shocks;
    \item \textbf{Interpretability:} Produces traceable geometric explanations aligned with real-world economic structure.
\end{enumerate}
This methodology equips downstream financial systems with calibrated, trust-aware recommendations under uncertainty.

\section{Experimental Setup}
We evaluate \textbf{RicciFlowRec} on a stress-aware financial asset recommendation task, assessing whether curvature-based root cause attribution enhances ranking robustness, interpretability, and shock resilience. Our evaluation builds on prior work in financial hypergraph modelling~\cite{harit2024breaking}, while introducing geometric causal tracing based on Ricci curvature flow.

\subsection{Data Sources}
Our analysis focuses on stocks listed in the S\&P 500 index\footnote{\url{https://www.kaggle.com/datasets/eli2022/yahoo-finance-apple-inc-gspc}}, filtered to retain 450 actively traded stocks from 2018 to 2023 with full price and volume history. Daily historical price and volume data are obtained from Yahoo Finance\footnote{\url{https://www.kaggle.com/datasets/deepakjoshi2k/yahoo-stock-prediction-by-news}}. Financial news sentiment is drawn from the Twitter Financial News Sentiment dataset\footnote{\url{https://huggingface.co/datasets/zeroshot/twitter-financial-news-sentiment}}, consisting of over 500,000 timestamped tweets annotated with positive, negative, or neutral sentiment labels.

\subsection{Data Alignment and Feature Fusion}
To construct a unified market view per trading day $t$, we synchronise all price and tweet data based on trading timestamps. For each stock $s_i$, we compute:
\begin{itemize}
    \item \textbf{Price features:} log returns, realised volatility (30-day rolling window), trading volume;
    \item \textbf{Sentiment features:} aggregated polarity scores from all tweets mentioning $s_i$ on day $t$.
\end{itemize}
A tweet is linked to a stock if it contains its ticker symbol (e.g., “\$APPLE”), company name, or industry keyword. Sentiment scores are computed via FinBERT~\cite{araci2019finbert}, and aggregated per stock per day using mean polarity. Stocks not mentioned in any tweet on day $t$ receive a neutral score (0).This alignment produces a day-wise feature vector $x_{i}^{(t)}$ for each stock $s_i$, combining both financial and news-derived information. All features are normalised using training-set statistics.

\subsection{Graph Construction}
We construct a dynamic financial graph $\mathcal{G}_t = (\mathcal{V}, \mathcal{E}_t)$ for each trading day $t$, where:
\begin{itemize}
    \item $\mathcal{V}$ includes stock nodes, macroeconomic indicators (e.g., bond yields, CPI), and named entities extracted from financial news using spaCy NER.
    \item $\mathcal{E}_t$ encodes market interactions based on:
    \begin{itemize}
        \item \textbf{Statistical links:} Rolling Pearson correlation (30-day window) of log returns;
        \item \textbf{Semantic links:} Cosine similarity between FinBERT\cite{araci2019finbert} embeddings of news content;
        \item \textbf{Causal structure:} Sector-level connections from financial knowledge graphs~\cite{harit2024breaking}.
    \end{itemize}
\end{itemize}
Edges are weighted and thresholded to retain only the top-k significant relations. Curvature is computed for all edges using Ollivier-Ricci geometry~\cite{ni2015ricci}.

\subsection{Stress Testing Protocol}
To evaluate model robustness, we inject synthetic volatility shocks into randomly selected nodes by increasing their realised volatility and negative sentiment scores. These shocks propagate through $\mathcal{G}_t$ based on curvature flow, allowing us to assess systemic risk attribution. This setup follows dynamic causal graph stress-testing protocols \cite{rojas2021stress}.


\subsection{Baselines}
We benchmark RicciFlowRec against five strong baselines:
\begin{itemize}
    \item \textbf{SS-GNN}~\cite{zhang2024self}: Self-supervised GNN using masked prediction and contrastive learning.
    \item \textbf{FinBERT-Ranker}~\cite{jiang2023financial}: Aggregates daily sentiment scores to rank assets.
    \item \textbf{GAT}~\cite{velickovic2017graph}: 2-layer Graph Attention Network.
    \item \textbf{CausalRec}~\cite{bonner2018causal,zhu2024deep}: Counterfactual recommender with inverse propensity weighting.
    \item \textbf{FinGNN}~\cite{lee2024stock}: Financial GNN integrating temporal and structural interactions.
\end{itemize}

All baselines use the same input data, train/val/test splits (with 2021–2022 as validation), and perturbation protocol. Hyperparameters are tuned via grid search using NDCG@10 on the validation set.

\subsection{Evaluation Metrics}
We report:
\begin{itemize}
    \item \textbf{NDCG@10}: Quality of top-ranked recommendations;
    \item \textbf{Top-10 Volatility}: Standard deviation of top-10 scores under input perturbations (lower is better);
    \item \textbf{RCA Fidelity}: Percentage of perturbed nodes successfully identified via curvature backtracking.
\end{itemize}

\subsection{Implementation \& Reproducibility}
Ricci curvature is computed using the \texttt{GraphRicciCurvature} package, with discrete-time Ricci flow simulated over 5-day windows. The GAT baseline uses 2 attention heads, 64 hidden units, and a dropout rate of 0.3. All models are trained using Adam (learning rate = 0.001) with early stopping.Experiments were conducted on Ubuntu~22.04~LTS with Python~3.11, using NetworkX~3.2 for graph construction, NumPy~1.26 for numerical operations, PyTorch~2.2 for learning components, Scikit-learn~1.5 for auxiliary modelling, and Pandas~2.2 for data handling. Computations were performed on an NVIDIA RTX~2080\,Ti GPU (11GB) with 64\,GB system RAM. A single run of the RicciFlowRec pipeline on a 252-trading-day rolling window required approximately 5\,minutes; inference updates completed in under 200\,ms. Hyperparameters were fixed across all runs unless stated otherwise:
\begin{itemize}
    \item Rolling window length $W$: 252 trading days
    \item Ricci curvature type: Forman
    \item Ricci flow iterations: 50
    \item Return forecast horizon $H$: 5 trading days
    \item Return–stability weight $\alpha$: 0.7
    \item RCA path maximum length $h_{\max}$: 6 hops
    \item Curvature change threshold $\theta$: -0.05
\end{itemize}


Data sources included daily S\&P 500 prices (2018–2023), macroeconomic indicators (e.g., CPI, treasury yields) and FinBERT-derived news sentiment. Preprocessing, including rolling window, correlation, sentiment embedding, and normalization, was applied consistently to RicciFlowRec and all baselines.

\subsection{RicciFlowRec Algorithm with RCA Path Extraction}
For each top-ranked asset $a_i$, we trace root cause analysis (RCA) paths in $\mathcal{G}_t$ via a \textbf{backward BFS} towards perturbed nodes, traversing only edges with $|\Delta \kappa| > \theta$. The search stops when a perturbed node is reached, the hop limit $h_{\max}$ is exceeded, or curvature decay falls below $\epsilon$. The path with the highest cumulative $|\Delta \kappa|$ is selected as the RCA path for $a_i$.

\begin{algorithm}[h]
\caption{RicciFlowRec Algorithm with RCA Path Extraction}
\KwIn{Time-indexed financial data $\mathcal{D}$, rolling window $W$, curvature threshold $\theta$, RCA max hops $h_{\max}$, Ricci flow iterations \textit{flow\_iters}, return–stability weight $\alpha$}
\KwOut{Ranked assets $\mathcal{R}$ with RCA paths $\mathcal{P}$}
\BlankLine
Initialize ranked list $\mathcal{R} \gets \emptyset$, RCA path set $\mathcal{P} \gets \emptyset$\;
\ForEach{time step $t$}{
    \tcp{1. Build dynamic financial graph}
    $G_t \gets$ build\_financial\_graph($\mathcal{D}_t, W$) \tcp*{Nodes: stocks, news, macro indicators; Edges: correlations, news similarity, economic links}
    
    \tcp{2. Compute Forman–Ricci curvature for each edge}
    \ForEach{$e \in E(G_t)$}{
        $\kappa[e] \gets$ compute\_forman\_ricci($e, G_t$)
    }
    
    \tcp{3. Simulate Ricci flow to propagate stress}
    $\kappa_{\text{flow}} \gets$ simulate\_ricci\_flow($\kappa$, \textit{flow\_iters})
    
    \tcp{4. Identify unstable zones}
    unstable\_nodes $\gets \{ v \in V(G_t) \mid \mathrm{avg\_curv\_change}(v) < \theta \}$
    
    \tcp{5. Score each asset}
    \ForEach{$a_i \in$ Assets($G_t$)}{
        risk $\gets$ compute\_risk\_exposure($a_i$, unstable\_nodes, $G_t$)\;
        score[$a_i$] $\gets \alpha \cdot \hat{r}(a_i) - (1-\alpha) \cdot \mathrm{risk}$\;
    }
    
    \tcp{6. RCA path extraction for interpretability}
    top\_assets $\gets$ select\_top\_k(score)\;
    \ForEach{$a_i \in$ top\_assets}{
        path $\gets$ backward\_search($a_i$, unstable\_nodes, $G_t$, $\theta$, $h_{\max}$)\;
        \tcp{Search only edges with $|\Delta\kappa| > \theta$; stop at $h_{\max}$ or decay $< \epsilon$}
        $\mathcal{P}[a_i] \gets$ path with maximum cumulative $|\Delta\kappa|$\;
    }
}
\Return{$\mathcal{R} \gets$ rank\_by\_score(score), $\mathcal{P}$}
\end{algorithm}

\section{Results and Discussion}
We report the performance of \textbf{RicciFlowRec} on the S\&P 500 dataset from 2018 to 2023, focusing on two core evaluation questions: (1) Does RicciFlowRec produce stable asset rankings under synthetic financial shocks? (2) Can it accurately identify the root causes of market disruptions through curvature-based attribution?

\subsection{Quantitative Performance}
To evaluate robustness under stress, we introduce synthetic volatility shocks and compare RicciFlowRec against five baselines using three metrics: NDCG@10 (ranking quality), Top-10 Volatility (stability), and RCA Fidelity (attribution accuracy).

\begin{table}[h]
\centering
\caption{Performance under synthetic volatility shocks (2021–2022). RCA Fidelity is only applicable to attribution-capable models.}
\label{tab:main-results}
\scriptsize
\begin{tabular}{lccc}
\toprule
\textbf{Model} & \textbf{NDCG@10} & \textbf{Top-10 Volatility} $\downarrow$ & \textbf{RCA Fidelity} \\
\midrule
FinBERT-Ranker & 0.421 & 0.083 & 0.00 \\
GAT & 0.445 & 0.077 & 0.00 \\
CausalRec & 0.458 & 0.064 & 0.58 \\
FinGNN & 0.471 & 0.069 & 0.62 \\
SS-GNN & 0.488 & 0.057 & 0.65 \\
\textbf{RicciFlowRec (ours)} & \textbf{0.512} & \textbf{0.041} & \textbf{0.78} \\
\bottomrule
\end{tabular}
\end{table}

RicciFlowRec outperforms all baselines across metrics. It achieves the highest NDCG@10 and the lowest volatility, indicating strong ranking quality and robustness. Its RCA Fidelity of 0.78 also surpasses other attribution-capable models, highlighting the effectiveness of geometric flow for causal reasoning.

\subsection{Qualitative Attribution}
To illustrate RicciFlowRec's root cause attribution, we inject shocks into financial entities and track curvature diffusion. RicciFlowRec identifies shock origins and traces their propagation across sectors. For instance, on 2022-03-08, negative sentiment on NVIDIA spread through the tech supply chain, enabling accurate attribution and re-ranking. This causal tracing aids real-time diagnosis of systemic vulnerabilities.

\begin{table}[h]
\centering
\caption{Root Cause Tracing via Curvature Flow. The RCA source is the top-ranked node in the curvature gradient path ($\Delta \kappa$).}
\label{tab:rca-examples}
\scriptsize
\begin{tabular}{lccc}
\toprule
\textbf{Date} & \textbf{Injected Perturbation} & \textbf{RCA Source} & \textbf{Impacted Sector} \\
\midrule
2022-03-08 & Semiconductor sentiment drop & NVIDIA & Tech supply chain \\
2021-12-15 & Energy price volatility & ExxonMobil & Manufacturing \\
2022-05-11 & Bond yield spike & US Treasury Node & Financials, REITs \\
2021-09-27 & Inflation concern & Procter \& Gamble & Retail, logistics \\
\bottomrule
\end{tabular}
\end{table}

\subsection{Ablation Study}
We conduct ablation experiments to evaluate the contribution of the key geometric components in RicciFlowRec: the Ricci curvature flow, which simulates the propagation of structural stress, and the RCA-aware ranking penalty, which adjusts scores based on curvature-derived risk attribution.

\begin{table}[h]
\centering
\small
\caption{Ablation of geometric components in RicciFlowRec.}
\label{tab:ablation}
\resizebox{0.48\textwidth}{!}{%
\begin{tabular}{lccc}
\toprule
\textbf{Model Variant} & \textbf{NDCG@10} $\uparrow$ & \textbf{Top-10 Volatility} $\downarrow$ & \textbf{RCA Fidelity} $\uparrow$ \\
\midrule
RicciFlowRec (full) & \textbf{0.512} & \textbf{0.041} & \textbf{0.780} \\
No Flow & 0.489 & 0.054 & 0.620 \\
No RCA Penalty & 0.494 & 0.049 & 0.440 \\
\bottomrule
\end{tabular}
}
\end{table}

Removing the Ricci flow reduces the fidelity of the attribution and the stability of the classification, showing the importance of modeling stress propagation. Excluding the RCA-aware penalty weakens the attribution performance, suggesting that curvature-based risk adjustment is critical for robustness and interpretability.

\subsubsection{Hyperparameter Sensitivity}
We evaluate the effect of the return stability weight $\alpha$ on RicciFlowRec performance. This parameter controls the trade-off between maximising expected return and minimising exposure to unstable regions in $\mathcal{G}_t$. We test $\alpha \in \{0.5, 0.7, 0.9\}$ and report AUC and RCA Fidelity.

\begin{table}[h]
\centering
\small
\caption{Effect of return–stability weight $\alpha$ on RicciFlowRec.}
\label{tab:alpha_sensitivity}
\begin{tabular}{c|cc}
\toprule
$\alpha$ & AUC $\uparrow$ & RCA Fidelity $\uparrow$ \\
\midrule
0.5 & 0.781 & 0.842 \\
0.7 & \textbf{0.794} & \textbf{0.857} \\
0.9 & 0.776 & 0.835 \\
\bottomrule
\end{tabular}
\end{table}

Performance is stable across the tested range, with $\alpha = 0.7$ achieving the best balance between return and stability.

\subsubsection{Curvature Threshold Sensitivity}
We assess the impact of the curvature change threshold $\theta$ used in RCA path extraction. This threshold determines which edges are considered unstable ($|\Delta\kappa| > \theta$). We evaluate $\theta \in \{-0.03, -0.05, -0.07\}$.

\begin{table}[h]
\centering
\small
\caption{Effect of curvature change threshold $\theta$ on RicciFlowRec.}
\label{tab:theta_sensitivity}
\begin{tabular}{c|cc}
\toprule
$\theta$ & AUC $\uparrow$ & RCA Fidelity $\uparrow$ \\
\midrule
-0.03 & 0.786 & 0.849 \\
-0.05 & \textbf{0.794} & \textbf{0.857} \\
-0.07 & 0.779 & 0.846 \\
\bottomrule
\end{tabular}
\end{table}
The default $\theta = -0.05$ yields the best overall performance, suggesting a balanced sensitivity to instability signals.

\section{Conclusion}

RicciFlowRec applies discrete Ricci curvature and Ricci flow to dynamic financial graphs to model shock propagation and enable root-cause analysis. By integrating geometric risk signals with return forecasts, it re-ranks assets for robustness and interpretability beyond sentiment-driven or static-graph baselines. On S\&P~500 data it delivers more stable rankings, clearer driver attribution and improved transparency under market perturbations. This geometric–causal view advances explainable finance and aids early warning and decisions. Future work includes causal discovery, richer macroeconomic inputs, and real-time deployment.






\bibliographystyle{ACM-Reference-Format}
\bibliography{sample-base}

\end{document}